\documentclass[letterpaper, 10 pt, conference]{ieeeconf}  

\IEEEoverridecommandlockouts                              

\usepackage[usenames] {color}
\definecolor{purple}{rgb}{0.4,0.2,0.8}

\long\def\ignorethis#1{}

\usepackage{algorithm}
\usepackage[noend]{algpseudocode}
\usepackage{graphicx}
\usepackage{subcaption}

\makeatletter
\def\BState{\State\hskip-\ALG@thistlm}
\makeatother

\newcommand{\pctab}{\hspace{0.2in}}

\usepackage{graphics} 
\usepackage{epstopdf}
\usepackage{epsfig} 
\usepackage{amsmath} 
\usepackage{amssymb}  
\usepackage{cite}
\usepackage{hyperref}
\usepackage{comment}

\title{\LARGE \bf
Feedback Control For Cassie With Deep Reinforcement Learning
}

\author{Zhaoming Xie$^{1}$ \and Glen Berseth$^{1}$ \and Patrick Clary$^{2}$  \and Jonathan Hurst$^{2}$ \and Michiel van de Panne$^{1}$
\thanks{$^{1}$Zhaoming Xie, Glen Berseth and Michiel van de Panne are with Department of Computer Science, University of British Columbia, Vancouver, BC, Canada. Email:
        {\tt\small \{zxie47, gberseth, van\}@cs.ubc.ca}. }%
\thanks{$^{2}$Patrick Clary and Jonathan Hurst are are with the Dynamic Robotics Laboratory, School of Mechanical, Industrial and Manufacturing Engineering, Oregon
State University, Corvallis, OR, USA. Email:{\tt\small \{claryp,jonathan.hurst\}@oregonstate.edu}}
}

\begin{document}

\maketitle
\thispagestyle{empty}
\pagestyle{empty}


\begin{abstract}
Bipedal locomotion skills are challenging to develop. Control strategies often 
use local linearization of the dynamics in conjunction with reduced-order abstractions to
yield tractable solutions.  In these model-based control strategies, the controller is often 
not fully aware of many details, including torque limits, joint limits, and other non-linearities 
that are necessarily excluded from the control computations for simplicity. 
Deep reinforcement learning (DRL) offers a promising model-free approach 
for controlling bipedal locomotion which can more fully exploit the dynamics. 
However, current results in the machine learning literature
are often based on ad-hoc simulation models that are not based on corresponding hardware.
Thus it remains unclear how well DRL will succeed on realizable bipedal robots.
In this paper, we demonstrate the effectiveness of DRL using a realistic model of Cassie, a bipedal robot.
By formulating a feedback control problem as finding the optimal policy for a Markov Decision Process, 
we are able to learn robust walking controllers that imitate a reference motion with DRL. 
Controllers for different walking speeds are learned by imitating simple time-scaled
versions of the original reference motion. Controller robustness is demonstrated through several challenging tests,
including sensory delay, walking blindly on irregular terrain and unexpected pushes at the pelvis. 
We also show we can interpolate between individual policies and that 
robustness can be improved with an interpolated policy.
\end{abstract}

\section{Introduction}
Walking dynamically like humans is a difficult task for bipedal robots due to inherent instability and underactuation.
Model-based approaches have demonstrated good success in this area, e.g., \cite{Posa16}, \cite{Da2016}. 
These approaches often require mathematical models to describe the robots dynamics to produce nominal trajectories in an offline process. 
Local feedback controllers are then designed to track these trajectories. 
However, the region of attraction for these controllers are often small, allowing for only modest perturbations. 
To overcome these, gait libraries\cite{Da2016} can be designed to increase robustness. 
Fundamentally, however, the design process and the inherent simplifications in designing the local feedback controllers may result in controllers that do not fully exploit the robot's capabilities. 
Deep reinforcement learning (DRL), on the other hand, provides a method to develop controllers in a model-free manner, albeit with its own learning inefficiencies. 
\cite{Peng2017} and \cite{Yu2018} have demonstrated that DRL can generate controllers for challenging locomotion skills. 
However, the articulated models used are simplified in nature and are not a representation of any particular existing bipedal robot. As a result, the prospects of deploying a learned controller on a large bipedal robot is not clear. 
While some work has shown the effectiveness of DRL on 
robots \cite{Sergey17}\cite{Peng18ICRA}, these robots are often fully actuated.

\begin{figure}%
\centering
\setlength{\fboxsep}{1pt}%
\begin{subfigure}{.24\textwidth}
  \includegraphics[width=\columnwidth]{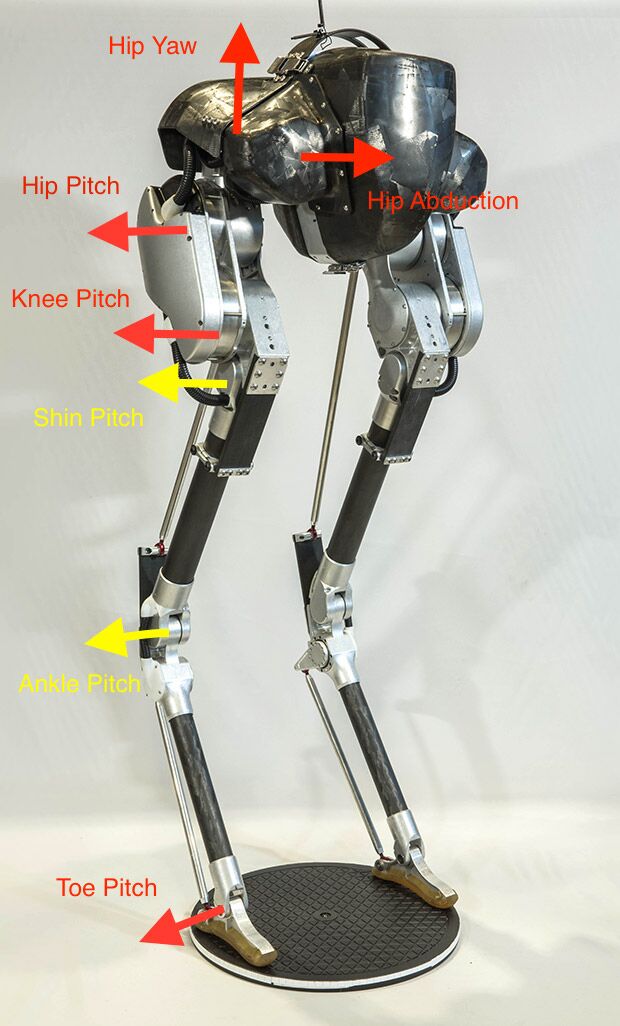}
\end{subfigure}%
\begin{subfigure}{.24\textwidth}
  \includegraphics[width=\columnwidth]{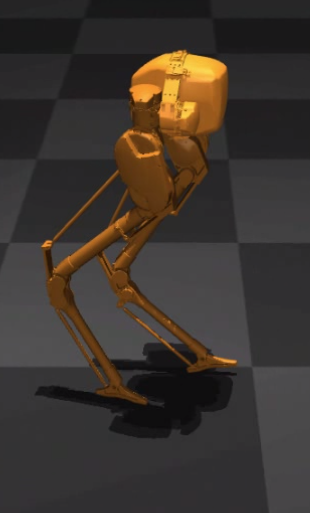}
\end{subfigure}%
\caption{Left: The bipedal robot Cassie developed by Agility Robotics. Each leg of Cassie has $7$ degrees of freedom, as indicated in the picture. Red arrows indicate active joints that can be controlled by motors and yellow arrows indicate passive ones. Right: Simulated model of Cassie in MuJoCo.}
\label{fig:cassie}
\end{figure}

This paper presents a framework for learning walking controllers on a realistic model of the Cassie biped, 
developed by Agility Robotics, shown in Fig.~\ref{fig:cassie}. 
By formulating a feedback control problem as searching for an optimal imitation policy for a Markov Decision Process, 
we can apply DRL to train controllers for bipedal walking tasks in a model-free manner with a single reference motion. 
Without needing to make the model-based simplifications commonly used to tractably realize control policies,
DRL is able to exploit the full dynamics of the robot and produce robust controllers. 
We test the robustness of our controllers by introducing sensory delays, 
testing blind walks on various types of terrain, and via random pushes applied to the body. 
Policies that can make the robot walk at a different speed can be constructed by 
retraining on a modified version of the reference trajectory that has been scaled in time. 
We can further make the robot speed up and slow down by interpolating between these policies. 
These results provide a degree of confidence that we can deploy a controller trained using DRL on a real biped.

\section{Related Work}
Deep reinforcement learning has been applied to multiple robotics tasks, e.g., \cite{Sergey17}\cite{Peng18ICRA}. 
While these results are impressive, the robots are fully actuated robotic arms. 
In this work, we are interested in the control of a bipedal robot to perform dynamic walking motions, 
which is inherently underactuated and unstable.

There exist various examples of using reinforcement learning for bipedal robot walking tasks, e.g. 
\cite{TedrakeLearningTW}\cite{Schuitema2010}. The work in this direction has often been applied to 
simple bipeds with large feet that will not fall during the training process\cite{TedrakeLearningTW} 
or that walk in 2D\cite{Schuitema2010}. By using a multi-layer neural network, 
we are able to train more capable controllers that can make an unstable biped walk in 3D.

The application of deep reinforcement learning to the problem of bipedal locomotion 
has recently become a popular benchmark problem in reinforcement learning. 
However, the actuators are typically assumed to be capable of large torques 
and the resulting motion is unrealistic \cite{Heess17} \cite{Merel17} \cite{Wang17}.
As a result, it is unlikely that these motions can be realized on a real robot. 
Even when the resulting motions are more realistic \cite{Peng2017} \cite{Yu2018}, 
the models considered are constructed from simple primitives such as boxes or cylinders, 
and are generally not good representations of real world robots. 
In this work, we directly work with a realistic model of the bipedal robot Cassie, 
where the actuators produce only limited torques that match the capability of the real robot, 
and physical constraints such as the reflected inertia of the actuators are also modeled in detail.

Feedback control methods have been shown to be successful on complex bipedal robots. 
Using the dynamics of the robot, desired behaviors can be computed offline using optimal control techniques 
such as direct collocation\cite{Hereid16} and quadratic programming (QP)\cite{Posa16}, 
and local control can then be computed online to make the robot track these desired motions. 
However, during the online control phase, linearization of the dynamics is often used, 
with a corresponding limitation on the robustness.
Gait libraries\cite{Da2016} can be used to help a robot handle larger perturbations 
by computing multiple gaits offline. In this work we use only a single reference trajectory 
and learn a controller that can handle large disturbances. 
Furthermore, since we are using model-free methods, the controller does not need to 
work with simplified models of the dynamics, as required by numerous model-based approaches.
Through experience, the full complexities and limitations of the robot dynamics can be taken into account 
by the controller. 
This comes with the caveat that the policy is developed in simulation and that additional effort may still be required to deal with the mismatch between the detailed simulation and reality.
\section{Background}
In this section, we briefly review reinforcement learning, feedback control and how we transform a feedback control problem into a reinforcement learning problem.
\subsection{Reinforcement Learning and Policy Gradient Methods}
A task in RL is often presented in as a Markov Decision Process (MDP) defined by a tuple $\{S, A, p, \gamma, r\}$, where $S \in \mathbb R^n$ is the state space, $A \in \mathbb R^m$ is the set of possible actions, and $\gamma \in [0, 1]$ is the discount factor. The transition function $p:S \times A \times S \rightarrow [0, 1]$ defines the dynamics of the system, and the reward function $r:S \times A \to \mathbb R$ provides the agent with a scalar reward at each state transition. A policy $\pi:S \times A \to [0,1]$ represents the action distribution given the current state $s_t$. The goal in reinforcement learning is to find a policy $\pi$ that maximizes the agent's expected return. In the case of a parametric policy $\pi_\theta$ with parameters $\theta$, the optimal policy can be determined by optimizing the objective $J(\theta)$,
\begin{align*}
\mathop{\mathrm{max}}_\theta J(\theta) = E_{\tau \sim \rho_{pi_\theta}(\tau) }\left[\sum_{t=0}^{T - 1}\gamma^t{r({s}_t, {a}_t)} \right]
\end{align*}
where $\rho_\pi$ represents the distribution of trajectories ${\tau = (s_0, a_0, ..., s_{T-1}, a_{T-1}, s_T)}$ induced by $\pi$.

In deep reinforcement learning, $\theta$ often represents the set of parameters for a multi-layer neural network. Policy gradient methods \cite{Sutton1999} are a popular class of algorithms for finding the optimal $\theta$. These methods iteratively improve the policy by estimating the gradient of $J(\theta)$ using rollouts and performing gradient ascent on the objective.

\subsection{Feedback Control}
In the control literature, bipedal robot locomotion control is often designed to be a feedback law for tracking some desired behavior. Given a dynamical system $x_{t+1} = f(x_t, u_t)$, where $x_t, x_{t+1} \in X \in \mathbb R^n$ are the states of the dynamical system at time $t$ and $t+1$, and $u_t \in U \in \mathbb R^m$ is the control input at time $t$, the equation of motion $f:\mathbb R^n \times \mathbb R^m \to \mathbb R^n$ describes how the dynamical system evolves over time. Trajectory optimization is often done offline to produce a nominal trajectory with $\hat{X}=\{x_0, x_1, \dots, x_T\}$ and $\hat{U}=\{u_0, u_1, \dots, u_{T-1}\}$that satisfies the equation of motion. Then a feedback law $u_t = g(x_t, \hat{x}_t)$ is calculated online to track the nominal trajectory by minimizing some distance metrics in $X$ and $U$. This usually involves solving a QP by linearizing the system dynamics along the nominal trajectory\cite{Hsu15}\cite{Posa16}. A popular choice is the Time Varying Linear Quadratic Regulator (TVLQR) \cite{tedrake2018}, where one solves the following QP: 
\begin{align*}
\text{minimize  }& \sum_{t=1}^{T-1} \delta_{u_t}^TR\delta_{u_t} + \delta_{x_t}^TQ\delta_{x_t} \\
\text{subject to  }& \delta_{x_{t+1}} =A_t\delta_{x_t} + B_t\delta_{u_t}
\end{align*}
Here $\delta_{x_t} = x_t-\hat{x}_t, \delta_{u_t} = u_t-\hat{u}_t$, and $A_t, B_t$ come from the linearized dynamics around $f(\hat{x}_t,\hat{u}_t)$. The distance metrics considered are quadratic functions defined in $X$ and $U$.

\subsection{Feedback Control interpreted as Reinforcement Learning Problem}
Given the dynamical system above and a reference motion $\hat{X}$, we can formulate an MDP. Let the set of states $S = X \times \hat{X}$, and actions $A = U$, then a natural choice for the transition function $p: S \times A \times S \to [0, 1]$ is by setting $p((f(x_t, u_t), \hat{x}_{t+1}) | (x_t, \hat{x}_t), u_t) = 1$ and setting it to 0 everywhere else. The reward function $r$ can be specified as the negative of the distance metrics defined in the space of $X$ and $U$. Then the original feedback control problem can be viewed as finding the optimal policy for this MDP, and can be solved using reinforcement learning algorithms.

Note that if we set the discount factor $\lambda = 1$, linearly approximate the transition function $p$ and let the metrics be quadratic functions, then finding the optimal policy is equivalent to solving a TVLQR. The result would be a time varying linear policy. But since it is based on a linear approximation, applying it to highly nonlinear systems like the Cassie robots can fail when subjected to large disturbance.

In this paper we use a reference trajectory with only state information, so we only consider metrics defined in the state space $X$ of the dynamical system for our reward function. By parameterizing the policy using a multi-layer neural network, therefore presenting a much richer policy space than that afforded by linear controllers. A more expressive policy may allow the controllers to develop more sophisticated recovery strategies in the presence of significant perturbations.

\section{methods}
\label{sec:methods}
\subsection{State Space and Action Space}
As mentioned in the previous section, the state for the MDP problem we are trying to solve is represented as the state and reference state of the dynamical system. In the case of Cassie, the state consists of the pelvis' position, orientation, velocity and angular velocity, plus the joint angle and joint velocity of all the active and passive joints. The combined representation yields an $82D$ state space. For this specific reference motion, the robot is moving in the $x$ direction, so we are neglecting the $x$ position in our state representation, yielding a final $80D$ state space.

\begin{figure}[h]
\centering
\includegraphics[width=8cm]{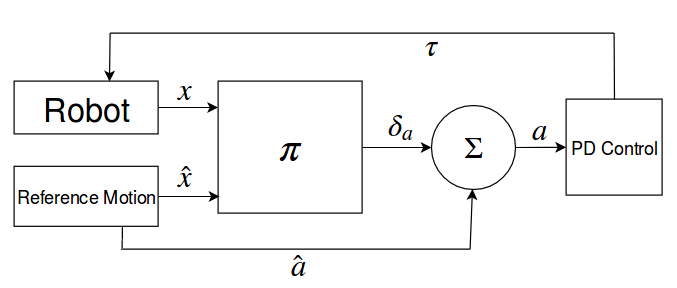}
\caption{Diagram for our control framework. The policy takes the robot state and reference state as input, and the output is added to the reference joint angles for the active joints. The result is then provided to a low level PD controller as target joint angles.}
\label{fig:framework}
\end{figure}

As shown by Peng et al.\cite{2017-SCA-action}, using PD control targets as the action space can greatly improve learning efficiency. We are choosing target joint angles for the active joints as our action $a$, and a low level PD controller applies the following torque to track these joint angles: $\tau = P(a-p_{active})-D(v_{active})$, where $P, D$ are the PD gains and $p_{active}, v_{acitve}$ are the current joint angles and joint velocities for the active joints. This produces a $10D$ action space. With the reference motion, a natural choice for target joint angles would be the reference joint angles. However, due to the underactuation of the robot, directly applying PD control to track the reference joint angles will quickly lead to the robot falling. However, the reference joint angles provide a hint as to what a desired action might be, instead of directly outputting the target angles, we choose to let our policy learn how to augment the reference angles. So, our policy output is in the form of $\delta_a$, and the action we use is $a = \hat{a} + \delta_a$, where $\hat{a}$ is the reference joint angles of the active joints from the reference motion. The control framework can be seen in Fig.~\ref{fig:framework}.

\subsection{Reference Motion and Simulation}
In principle, our framework can incorporate any reference motion that describes how the robot is expected to move over time. In this paper, we use a reference motion of Cassie taking two steps in the $+x$ direction. We then extend this two-step motion by copying it repeatedly, except for the value describing the $x$ position of the pelvis, which increases smoothly over time to represent the robot's continuous forward movement.

At the beginning of each episode, the pose of the robot is set to a state randomly selected from the reference motion.
To avoid the excess exploring of poor states, we terminate the episode early and set the remaining rewards to be $0$ if the reward the robot received in the previous state is below some value or when the height of the pelvis makes the robot unstable. Further details are described in Section V.

\subsection{Network and Learning Algorithm}
\label{sec:methods-net-and-learning-alorithm}
\begin{figure}[h]
\centering
\includegraphics[width=8cm]{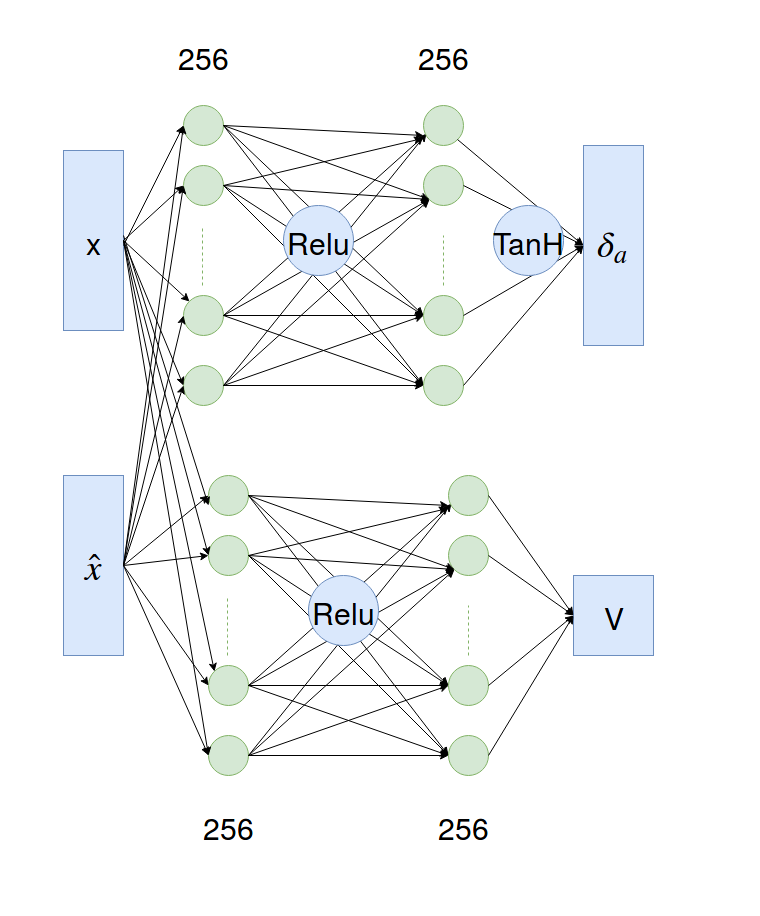}
\caption{The actor-critic network used to parametrize the policy and the value function. The output of the policy network $\delta_a$ is added to the current reference angle of the active joints. The result is then used as the target joint angles for PD control.}
\label{fig:actor_critic}
\end{figure}

We use an actor-critic learning framework for our experiments. The actor and critic are parametrized by neural networks with parameters $\theta$ and $\phi$, the network structure is shown in Fig. \ref{fig:actor_critic}. Two-layer neural networks are used for both actor and critic, with all hidden layers having a size of $256$. ReLU activations are used between the hidden layers, and the output of the actor is passed through a $TanH$ function to limit the range of the final output.

We use the Proximal Policy Optimization (PPO) \cite{SchulmanWDRK17} to optimize our policy. PPO is an on-policy, model-free reinforcement algorithm based on the stochastic policy gradient framework, and there exist multiple variations in practice.
PPO has been successfully applied to bipedal locomotion tasks~\cite{Heess17}\cite{Yu2018}.
As pointed out by Henderson et al.~\cite{Henderson2017}, the implementation details can significantly impact the performance of reinforcement learning algorithms, and thus here we briefly describe our implementation.

Before each network update, we sample $N$ trajectories of maximum length $T$ using the current policy, each trajectory starts from a random pose sampled from the reference motion. Tuples of the transitions and rewards are collected as $\{s_t, a_t, s_{t+1}, r_t\}$, where $s_t = (x_t, \hat{x}_t)$ and $a_t=u_t$ are the state and action at time $t$ from a trajectory.
The value of $s_t$ is estimated using these samples,
\begin{align*}
\hat{V}_t=\sum_{t'=t}^T \gamma^{t'-t}r_{t'}+V_\phi(s_{T+1}).
\end{align*}
The critic network parameters $\phi$ are updated by minimizing:
\begin{align*}
L_{V}(\phi) = \frac{1}{N}\sum_{i=1}^N(\frac{1}{T}\sum_{t=1}^T (\hat{V}^i-V_\phi(s^i_t))^2).
\end{align*}
where the superscript $i$ indicates the $i$th trajectory sampled.

The actor network is updated using the estimated advantage function:
\begin{align*}
\hat{A}_t = \hat{V}_t-V_{\phi}(s_t).
\end{align*}
Let $\pi_{old} \leftarrow \pi_\theta$ and $\rho_t =\frac{\pi_\theta(a_t | s_t)}{\pi_{old}(a_t | s_t)}$. We update the parameters $\theta$ of the actor network by maximizing the following objective:
\begin{align*}
L_{ppo}(\theta) = \frac{1}{N}\sum_{i=1}^N\frac{1}{T}\sum_{t=1}^T
 \min(\rho_t^i\hat{A}^i_t,clip(\rho^i_t,1-\epsilon,1+\epsilon)\hat{A}^i_t).
\end{align*}
where $clip(c, a, b) = c$ if $c\in [a, b]$, or $c=a$ if $c<a$, and $c=b$ if $c>b$. $\epsilon$ is chosen to be $0.2$ in our implementation.

\section{Results}

The results are illustrated in the associated video\footnote{\url{https://www.youtube.com/watch?v=z3DMKQwt68Y}}.
We evaluate our method using a simulated model\footnote{\url{https://github.com/osudrl/cassie-mujoco-sim}} 
of Cassie in the MuJoCo\cite{mujoco} simulation environment. Cassie is a bipedal robot developed by Agility Robotics (Fig.~\ref{fig:cassie}). It has $20$~degrees of freedom and $10$~actuators. In addition to the $6$~unactuated degrees of freedom for the floating base, each leg of Cassie has unactuated spring joints, resulting in a challenging control problem. A reference controller was implemented using manually-tuned heuristics for producing stable walking, and has its performance compared to that of our learned feedback controller in a later section. A walking reference trajectory is created from this reference controller for Cassie. The reference trajectory contains two full footsteps of data, in which Cassie moves forward $0.5$ meters from its original location in about $0.7$~seconds, sampled at $32$~ms intervals. We implement our neural network using Pytorch\cite{paszke2017automatic}, and the experiments are run on a eight-core computer using a single thread. During training, the simulation rate is set to $1$~kHz to lower the computational requirements, while during testing the simulation is run at $2$~kHz to match the control rate of the real robot. Target joint angles are computed every $32$~ms, giving us a policy query rate of $31.25$~Hz, while the low level PD controller is run at the same rate as the simulation.

We first collect $50,000$ states by sampling trajectories starting from random poses in the reference trajectory using a random policy. 
The mean and standard deviation is computed from the collected samples. These values are then used to normalize the inputs during training, similar to batch norm.

During training, we sample the action using a Gaussian policy to encourage exploration, with the policy mean being the outputs of the actor network and the covariance $\Sigma$ being diagonal matrix with diagonal elements set to $0.018$. In test time, we directly use the output of the actor network as the policy.

The reward function used is defined as
\begin{equation*}
\begin{split}
r = w_{joint}r_{joint}+w_{rp}r_{rp}+w_{ro}r_{ro}+w_{spring}r_{spring},
\end{split}
\end{equation*}
where $r_{joint}$ measures how similar the active joint angles are to the reference active joint angles, $r_{rp}, r_{ro}$ measures how similar the pelvis position and orientation are to the reference motion and $r_{spring}$ is an additional term to help stabilize the springs on the shin joints. 
We currently set the weights from experience without any special efforts to fine-tune it.
The joint differences are computed as $r_{joint} = \exp(-||x_{joint}-x_{refjoint}||^2)$. The rest of the terms are computed similarly, and the weights are $0.5, 0.3, 0.1, 0.1$, respectively. Note that the reward is in the range of $[0,1]$. We ignore the passive spring-loaded joints in our reward calculation because the deflections are relatively small due to the spring stiffness.

Episodes are stopped when a termination condition is met or when they reach the maximum length $T$, which is set to $300$ control steps ($\sim 10$ seconds) in all experiments. 
The termination condition we use is whether the height of the pelvis is lower than $0.6$ meter or higher than $1.2$ meters, or the reward is less than $0.6$.
We collect a maximum of $3,000$ samples. 
These samples are then used for updating the actor and critic networks. 
The update is done by performing stochastic subgradient descent on the loss functions described in Sec.\ref{sec:methods-net-and-learning-alorithm}. 
We use a batch size of $128$ and perform $64$ updates using the Adam optimizer\cite{Kingma14}. The step size for Adam is initially set to $1e^{-3}$ for the actor network, and $1e^{-2}$ for the critic. These are then decreased by $1\%$ after each iteration, until the step size is less than $1e^{-4}$ for actor and $1e^{-3}$ for critic.

\subsection{3D Walking}
\begin{figure}[tb]
\includegraphics[width=\columnwidth]{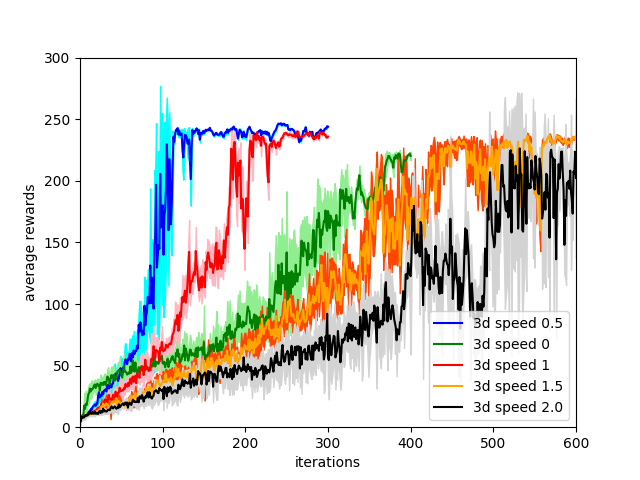}
\caption{Learning curves for training with different target movement speeds. The speed is a function of the distance the reference motion travels in a single stride (i.e. two steps). The reference trajectory is $0.5$ meter per stride.
}
\label{fig:learningcurve}
\end{figure}

We use our framework to train a controller of the 3D simulated model of Cassie using a reference motion. The resulting learning curve is shown in Fig.~\ref{fig:learningcurve}. The policy reaches peak performance in the first $200$ iterations, and we stop training after $300$ iterations, which takes around $2.5$ hours.
Please see the accompanying video for more results. 
Without external disturbances, the learned controller can accumulate a total reward of about $250$ 
where the maximum possible reward is $300$. 

\subsection{Sensory Delay}
We simulate sensory delay by feeding the robot state information from a few milliseconds in the past. The controllers perform well with a $5$~ms delay, accumulating a reward of 247 on average over 10~runs.
However the robot quickly falls with a delay of $10$~ms, although we are also assuming
that the delay impacts the PD-control loops.

\subsection{Terrain and Perturbation Test}
We further evaluate the robustness of our learned controller by having the robot walk blindly across uneven terrain. The terrain is generated in Mujoco using a sinusoidal function. The height map is in the form $z=h\sin{x}$, where $z$ is the height of the terrain, $x$ is the distance in meters along the direction the robot is heading, and $h$ is the center-to-peak  terrain height ratio. 
We compare the robustness of our learned controller to the reference controller on terrain with different $h$ values. The reference controller can walk across sinusoidal terrain with $h=0.07$ without falling, while our learned controller can handle up to $h=0.15$ without falling. 
For the perturbation test, our learned controller can recover from pushes that last for $0.2$~s with a magnitude of $140$~N in the forward direction, $90$~N in the backward direction, and $50$~N from either the left or right side. In comparison, the reference controller can cope with disturbances of up to $50$~N in the forward direction, $70$~N in the backward direction, and $15$~N to the sides.

\subsection{Different Speed}
To see if we could learn feedback controllers of different speeds, we train different policy for different reference motion by stretching and compressing the x position of the pelvis in the original reference motion.
Note that by varying the speed in this way, the resulting reference motion is no longer physically feasible, with the foot sliding along the ground when it should be fixed. 
We compute solutions for speeds of $0,2,3,$ and $4\times$ the original reference motion. 
The training for each speed takes $3$ to $5$ hours, and the system can learn controllers that successfully track the desired speed with reasonable motion without any additional hyperparameter tuning. Fig.~\ref{fig:learningcurve} shows the learning curves for training with reference motions of different speeds.  
Note that the learning process has become unstable for reference motion with a $4\times$ speed.
We believe this result stems from the policy making tradeoffs between following a physically infeasible motion and tracking the desired speed.

\subsection{Interpolation}
We explore the possibility of having Cassie speed up and slow down by interpolating between different policies trained with different target speeds. Given an interpolation parameter $\lambda \in [0, 1]$, and policy $\pi_1, \pi_2$ with reference motion $\hat{X}_1, \hat{X}_2$, we construct a interpolated reference motion as $\hat{X} = \{\hat{x}_t = \lambda \hat{x}_{1t} + (1-\lambda)\hat{x}_{2t} | \hat{x}_{1t}\in\hat{X}_1, \hat{x}_{2t}\in\hat{X}_2 \}$, and given $\hat{x}_t \in \hat{X}$, the interpolated policy would be $\pi(a |(x, \hat{x}_t)) = \lambda\pi_1(a | (x, \hat{x}_{1t}))+(1-\lambda)\pi_2(a |(x,\hat{x}_{2t}))$. 

With this interpolation scheme, we let $\lambda = 1-0.625t_{sim}$ until $\lambda$ is equal to $0$, where $t_{sim}$ is the clock in simulation. We can successfully speed up and slow down the gait by interpolating between a policy that 
walks $0.5$~m per stride and $1.0$~m per stride, as well as between $1.0$~m per stride and $1.5$~m per stride.

We also test if interpolating in this way can increase the robot's robustness on uneven terrain, which is similar to the gait library used in~\cite{Da2016}. We let the robot interpolate between $0.5$~m per stride and $1.0$~m per stride based on the current speed of the pelvis, which naturally speeds up and slows down when going up and down slopes. With an interpolated policy, the robot can handle sinusoidal terrain of $h = 0.22$, a significant increase over the non-interpolated case. 

\section{Conclusion}
We have presented a framework for feedback control of bipedal walking that imitates a given reference trajectory. 
By parameterizing the policy using a multi-layer neural network and applying policy-gradient learning,
we are able to learn controllers that are robust to large disturbances. 
We show that our framework can produce reasonable controllers even with physically infeasible reference trajectories,
such as those resulting from simple retiming of the reference motion. 
We investigate the robustness of the learned controllers by modeling sensory delay, 
evaluating blind walking across varying terrain, and applying large pushes to the pelvis while walking. 
We further demonstrate the ability to interpolate between policies that represent different walking speeds, 
and show that using an adaptive interpolated policy yields more robust gaits.

Our controllers still rely on full state information from the robot, 
while in a real world scenario the state must be estimated from noisy sensor measurements. 
We are currently extending our framework to work directly with output (sensory) feedback. 
Furthermore, even though the learned blind controllers can robustly handle unexpected terrain, 
we believe robustness can be greatly improved by incorporating terrain knowledge, i.e., via vision. 

In principle, diverse behaviors can also be obtained via additional reference motions, 
as we demonstrate by varying the speed of the reference motion. 
We plan to generate further behaviors such as jumping and running
with the help of reference motions, which could be designed or computed in various ways.
Another future direction would be to learn a unified controller that can perform feedback control 
given any reasonable reference motion, to allow for zero-shot learning when given new motions for new tasks.

All our experiments are currently performed in a simulated environment. 
As a next step, we plan to evaluate the resulting policies on the real robot, 
possibly with on-the-robot fine-tuning to cross the reality gap.
\section*{Acknowledgements}
We thank Xue Bin Peng and Xiaobin Xiong for helpful discussions. This work was funded in part by NSERC Discovery RGPIN-2015-04843.
\addtolength{\textheight}{-9.50cm}   
\bibliographystyle{IEEEtran}
\bibliography{iros}

\end{document}